\documentclass[letterpaper]{article}
\usepackage{natbib,alifeconf}
\usepackage{amsmath}
\usepackage{graphicx,wrapfig,hyperref}

\usepackage{algorithm}
\usepackage{algpseudocode}
\floatname{algorithm}{Procedure}

\usepackage{siunitx}
\usepackage{hyperref}

\newcommand{\vect}[1]{\boldsymbol{#1}}

\newcommand{\fig}{Fig.~}

\title{Locally adaptive cellular automata for goal-oriented self-organization}

\author{Sina Khajehabdollahi$^{1}$, Emmanouil Giannakakis$^{1, 2}$, Victor Buend\'ia$^{1,2}$, Georg Martius$^4$, Anna Levina$^{1, 2, 3}$ \\
\mbox{}\\ 
$^1$Department of Computer Science, University of T\"ubingen, T\"ubingen, Germany\\
$^2$Max Planck Institute for Biological Cybernetics, T\"ubingen, Germany\\
$^3$Bernstein Center for Computational Neuroscience  T\"ubingen, T\"ubingen, Germany\\
$^4$ Max Planck Institute for Intelligent Systems, T\"ubingen, Germany\\
sina.abdollahi@gmail.com \\
\bigskip
}

\begin{document}
\maketitle

\begin{abstract}
The essential ingredient for studying the phenomena of emergence is the ability to generate and manipulate emergent systems that span large scales. Cellular automata are the model class particularly known for their effective scalability but are also typically constrained by fixed local rules.  In this paper, we propose a new model class of adaptive cellular automata that allows for the generation of scalable and expressive models. We show how to implement computation-effective adaptation by coupling the update rule of the cellular automaton with itself and the system state in a localized way. To demonstrate the applications of this approach, we implement two different emergent models: a self-organizing Ising model and two types of plastic neural networks, a rate and spiking model. With the Ising model, we show how coupling local/global temperatures to local/global measurements can tune the model to stay in the vicinity of the critical temperature. With the neural models, we reproduce a classical balanced state in large recurrent neuronal networks with excitatory and inhibitory neurons and various plasticity mechanisms.  
Our study opens multiple directions for studying collective behavior and emergence.

\end{abstract}

\section{Introduction}

Cellular automata (CA) are simple models of computation where cells organized in a regular grid update their state according to rules that are local functions of the cell's neighborhood \citep{neumann1966theory, chopard1998cellular}. It has long been observed that CAs can exhibit complex pattern formation and highly non-trivial dynamics. The simplicity of the basic concept, combined with their extreme versatility make CAs a great tool for studying the phenomena of emergence and complexity in a mathematical framework \citep{wolfram1984cellular}. CAs have been used to model a variety of physical \citep{vichniac1984simulating, toffoli1984cellular}, biological \citep{langton1986studying, turing1952chemical, mordvintsev2020growing, Farner2021}, and more recently, differentiable, goal-oriented artificial intelligence phenomena \citep{chan2020lenia, mordvintsev2020growing, variengien2021towards, randazzo2020self-classifying, Pontes-Filho2022}.

The success of CAs in modeling complex physical and biological phenomena is perhaps in no small part due to the model's inherent inductive bias that reflects reality reasonably well. Any CA model is inherently local and computes and distributes information locally, a constraint that also exists strongly in the real world \textit{(ignoring spooky action at a distance)}. Due to this constraint, CAs optimized for certain problems will always solve them via collective local interaction. CAs, therefore, allow us to study how self-organizing systems with only local interactions can give rise to global structure and complexity, a key characteristic of living matter.

However, for the very reason they are efficient, the expressive capabilities of CAs are constrained by their design principles. Namely, the homogeneity of the update rule in space and time forces every cell in the system to update according to the same rule. This means that CA models, which can be thought of as mesoscopic/coarse-grained models of phenomena, are fixed in the level of abstraction that they are modeling. Some work on multi-scale cellular automata has shown the capacity of such models to generate multi-scale patterns when multiple rules are interacting or competing \citep{mccabe2010cyclic, rampe2011, slackermanz2021}. These systems demonstrate how context-specific heterogeneity in the application of these CA rules can allow for another qualitative change in the emergent dynamics. This multi-scale property may, in fact, be crucial to the top-down organizational properties of biology \citep{pezzulo2016top}. Most existing models still only have a fixed set of update rules that are then applied selectively and cannot be flexibly modified as in nature. There are variations of CAs that address this homogeneity in different ways, such as probabilistic CAs \citep{louis2018probabilistic} or models of neural cellular automata manifolds \citep{hernandez2021neural}. Here, we present a more general description of adaptive CAs that can be extended to a variety of domains.

Unfortunately, much of the CAs' utility comes from their computational efficiency and simplicity. Implementing heterogeneity or adaptation in CAs can make them too computationally expensive at the scales they require to become interesting. Here, we propose a method for designing adaptive cellular automata that minimizes the impact of adaptation on computational efficiency (\fig \ref{fig:benchmark}) by taking advantage of advances in computational power. In particular, with modern GPU hardware and fast linear algebra libraries it becomes very cheap to write our CAs as highly parallelizable matrix operations such as convolutions. In turn, this allows for a massive increase in the scale, expressivity, interactivness and ability to train/differentiate such models to the extent that they can be run on personal computers with high frame rates.

The methods discussed in this paper are meant as building blocks for designing models of collective systems that exhibit emergence. We follow here the principle of building intuitions about such systems through the process of constructing them \citep{neumann1966theory}. This will then further allow us to study these systems in quantitative ways, build better models, and guide our theories. Having toy models of emergence that simulate or approximate the levels of emergence we observe in nature is a crucial step toward expanding our theories of complexity and self-organization. To that end, we implement as adaptive CAs three known foundational models of collective dynamics, the 2D Ising model, a Wilson-Cowan rate neural network, and a leaky integrate-and-fire spiking neural network. As these models can often be generalized across domains as generic models of interacting systems, our aim is to explore methods of local adaptation that can self-organize these general systems towards desirable macroscopic states.

\section{Method}
\begin{figure}[!htb]
\begin{center}
\includegraphics[width=8cm]{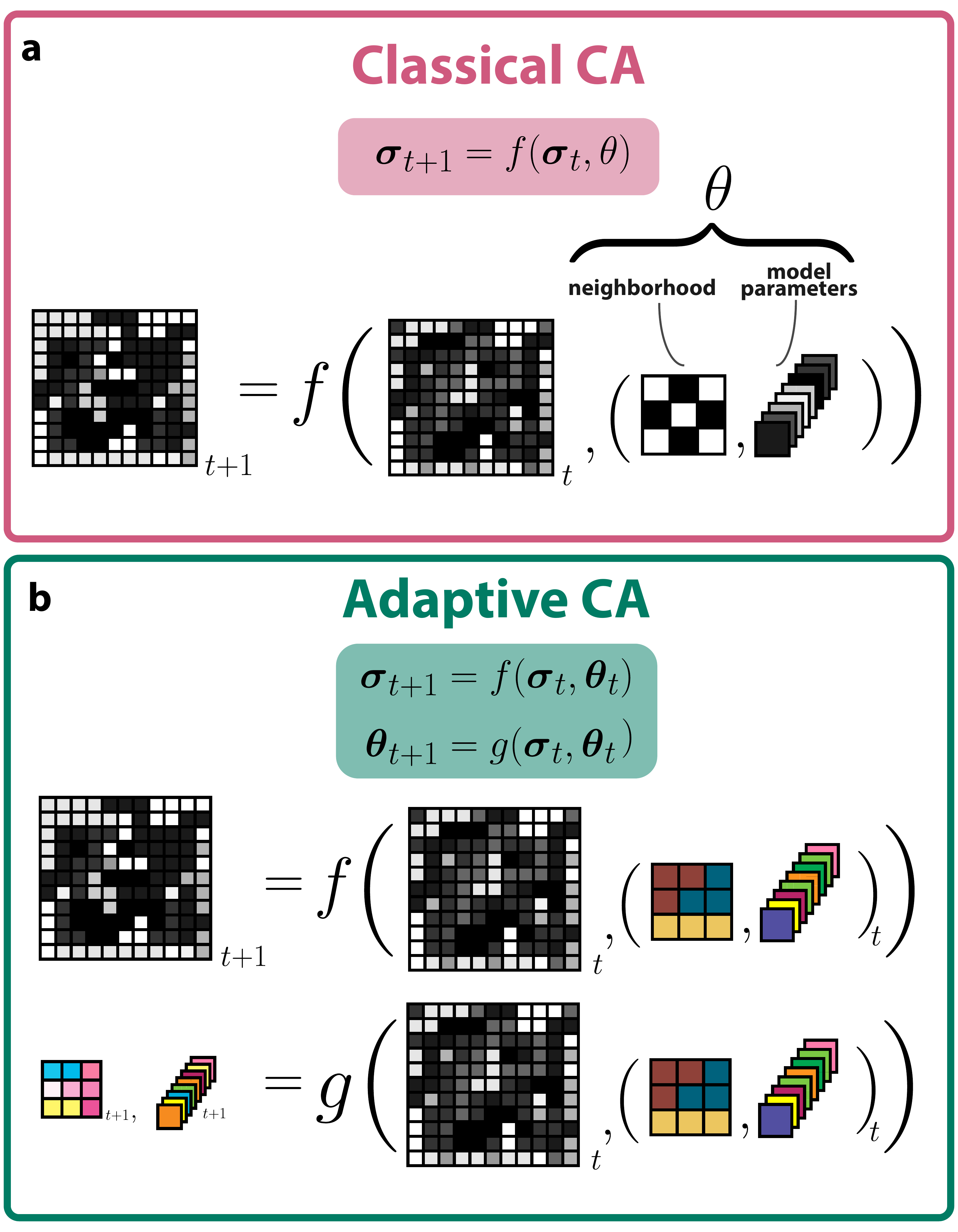}
\caption{\textit{(\textbf{a}): Classical CAs are dynamical systems parameterized by fixed update rules that are applied identically at every cell in the grid. (\textbf{b}): Adaptive CAs extend the same principles of classical CAs onto the update rule itself, allowing the rule to also change in space and time. The evolution of the update rule can itself behave like a CA, reacting to its neighbors' activities and patterns. Coupling these systems allows one to create and study much more expressive models that simulate complex phenomena such as driven, dissipating, and self-organizing non-equilibrium dynamics that are characteristic of life.}}
\label{fig1}
\end{center}
\end{figure}

\begin{figure}[!htb]
\begin{center}
\includegraphics[width=8cm]{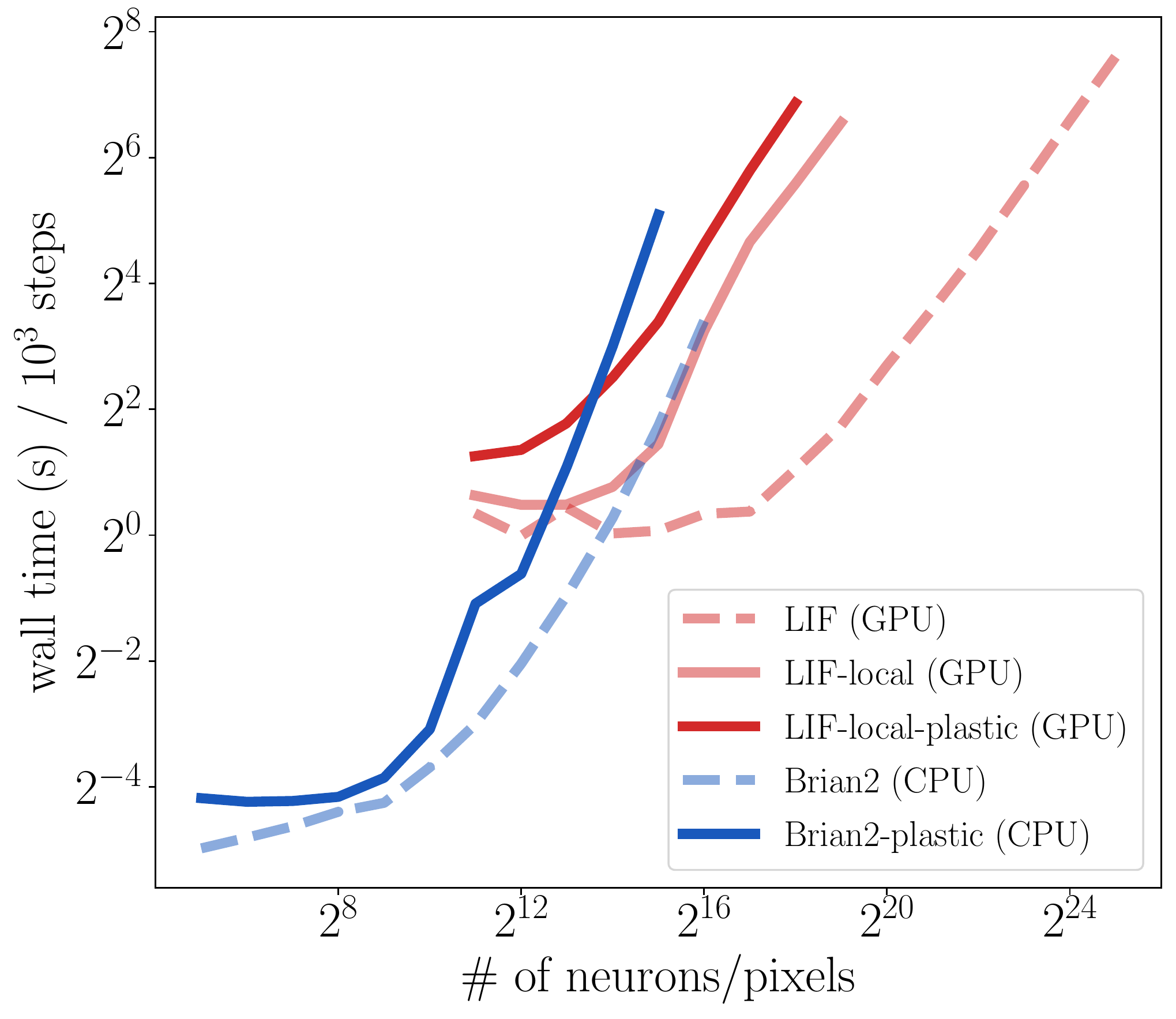}
\caption{\textit{Benchmark of our leaky integrate-and-fire adaptive CA model with homogeneous kernels (LIF), heterogeneous kernels (LIF-local), and plastic kernels (LIF-local-plastic) running on a single 12GB GTX 2080 compared to neural network simulations run with Brian2 (with and without plasticity) on a 8-core CPU with a single thread. All models were scaled until failure/could not be run on a single computer.
}}
\label{fig:benchmark}
\end{center}
\end{figure}

Update rules for cellular automata are traditionally fixed and applied identically and in parallel at every cell on the grid. By reusing a fixed rule that is always applied identically, CAs can take advantage of highly parallelizable algorithms with minimal memory costs. These update rules can often be written as sequences of matrix multiplication operations, which when implemented with modern linear algebra libraries and GPUs offer very efficiently computed models that scale well. In this framework, convolutions are a natural choice to accomplish the types of local calculations required by cellular automata \citep{gilpin2019cellular, mordvintsev2020growing}. However, 2D convolutions (Procedure \ref{alg:conv2d}) use a fixed kernel, which means that a CA that makes use of these operations will also have a fixed update rule at all times, see \fig\ref{fig1}a. However, with the relative growth of GPU memory capacity and computational power, it has become increasingly feasible to embed localized rule parameters in memory and add a new dimension of heterogeneity to these models \citep{hernandez2021neural}.

To allow cellular automata to have \textit{adaptive} update rules that can change flexibly as the system evolves in time, we define recursive update rules where the rule parameters are embedded into the state of the system, see \fig\ref{fig1}b (Procedure \ref{alg:sliding_window}). To do this, we can concatenate the state $\vect{\sigma}_{i}(t)$ of a cell with the parameters  $\vect{\theta}_{i}(t)$ of a local update rule, such that the state and update rule of a cell at time $t$ is fully defined by
\begin{equation}
    \vect{s}_{i}(t) = [\vect{\sigma}_{i}(t), \vect{\theta}_{i}(t)].
\end{equation}
Whereas in a classical CA, the update rule $f(\vect{\sigma}_{i}(t))$ would be fixed, with adaptive CAs, the update rule 

\begin{equation}
    \vect{s}_{i}(t+1) = f(\vect{s}_{i}(t)) = f(\vect{\sigma}_{i}(t), \vect{\theta}_{i}(t))
\end{equation}
is local and parameterized by $\vect{\theta}_{i}(t)$.
There are a number of different ways one can parameterize a local update rule, for example, the weights of the forces a cell feels from its neighbors, a decision tree of a set of locally context-dependent update rules, and so on. The toy models shown here are all implemented using the PyTorch library \citep{NEURIPS2019_9015}.

\begin{algorithm}
\caption{2D Convolution}\label{alg:conv2d}
    \begin{algorithmic}[1]
        \Function{conv2d}{$image, kernel$}
            \State $patches \gets$ \texttt{unfold($image, patchSize$)}
            \For{$p_{ij}$ in $patches$}
                \State $out_{ij} \gets$ \texttt{sum($p_{ij} \odot kernel$)}
            \EndFor
            \State \textbf{return} $out$
        \EndFunction
    \end{algorithmic}
\end{algorithm}

\begin{algorithm}
\caption{Generalized sliding-window function}\label{alg:sliding_window}
    \begin{algorithmic}[1]
        \Function{local}{$image, \theta$}
            \State $patches \gets$ \texttt{unfold($image, patchSize$)}
            \State $\Theta \gets$ \texttt{unfold($image, patchSize$)}
            \For{$p_{ij}, \theta_{ij}$ in \texttt{zip}$(patches, \Theta)$}
                \State $out_{ij} \gets$ \texttt{adaptive\_rule}$(\theta_{ij})(p_{ij})$
            \EndFor
            \State \textbf{return} $out$
        \EndFunction
    \end{algorithmic}
\end{algorithm}

Ultimately, there is still a bedrock level in which the model is fixed that is determined by the nature of the update rule's own update rule. However, by making the update rule adaptive, we can allow CAs to navigate through a much larger space of dynamical systems, giving it the possibility to navigate its phase space according to some driving principle or goal.

\begin{figure*}[!ht]
\centering
\includegraphics[width=\textwidth]{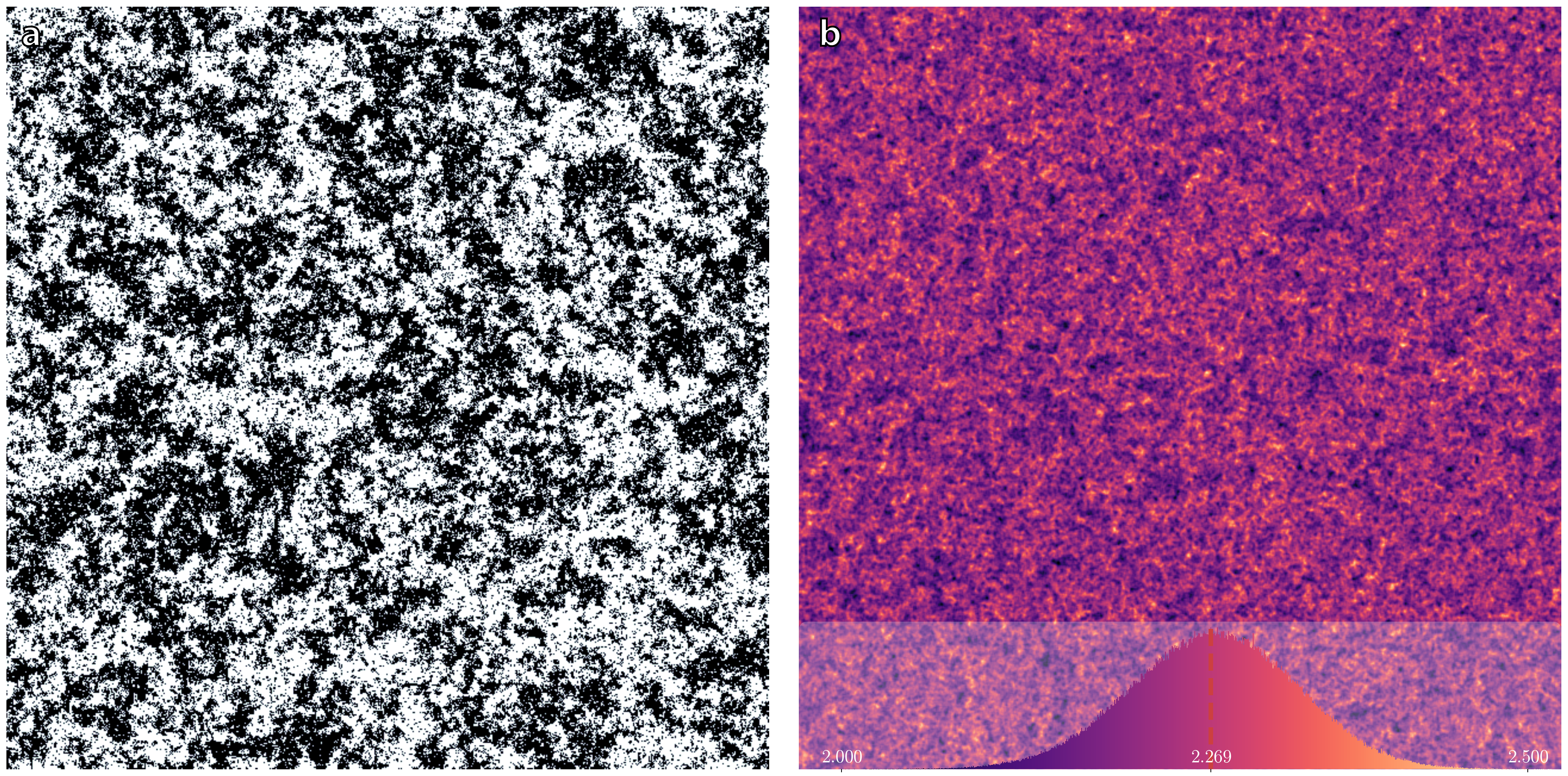}\vspace{-0.5em}
\caption{\textit{Spins (\textbf{a}) and temperatures (\textbf{b}) of a local adaptive Ising model of size 1280x1280 near $T_c \approx 2.269$. Overlay in (b) shows the distribution of local temperatures (centered at critical temperature) and indicates the color code}.}
\label{isingCA_hires}
\end{figure*}

\begin{figure*}[!ht]
\centering
\includegraphics[width=\textwidth]{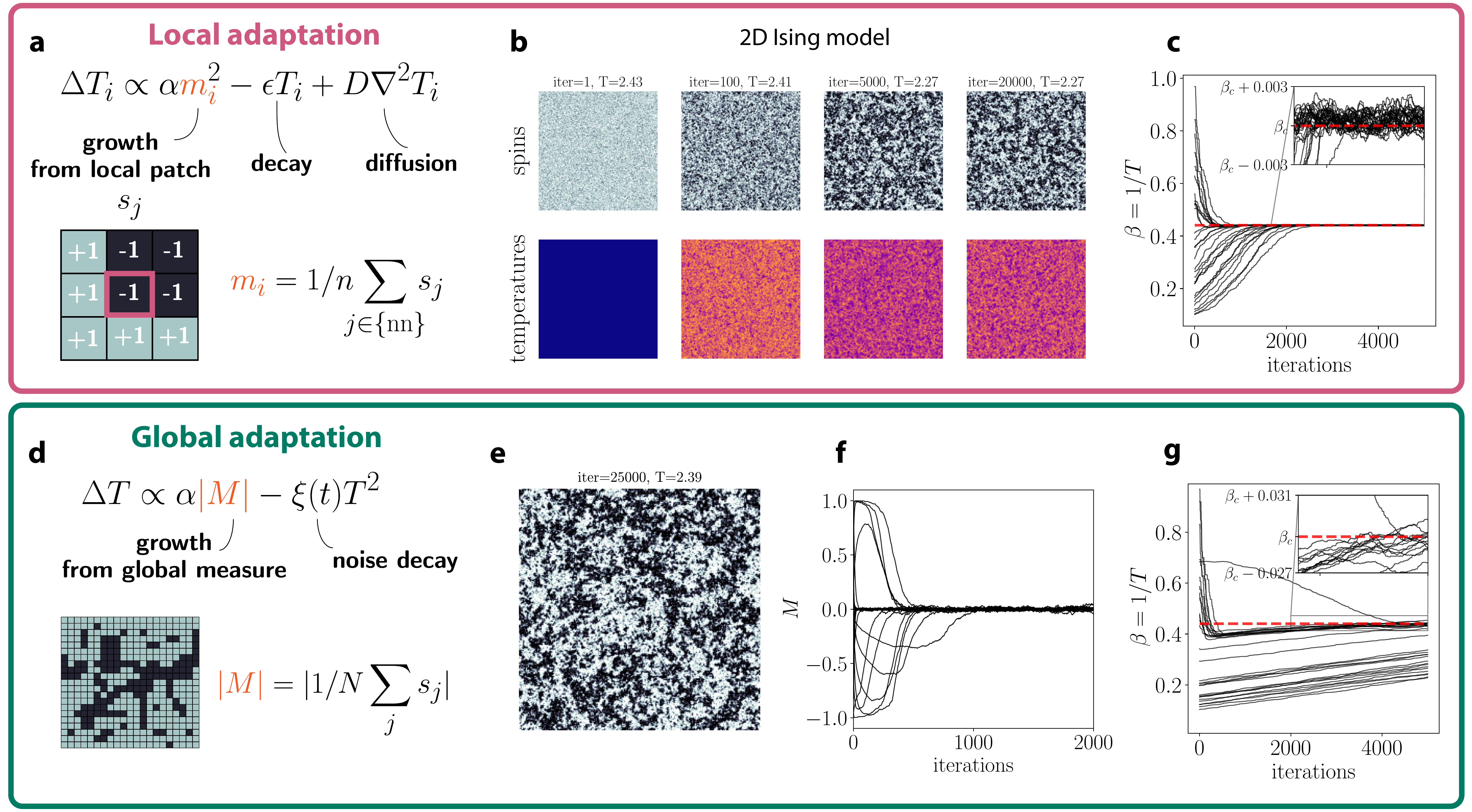}\vspace{-0.5em}
\caption{\textit{Two methods to achieve self-organization to criticality in the 2D Ising model: a local/microscopic method (top panels, local temperature evolution described in \textbf{a}), and a global/macroscopic method (bottom panels, global temperature update in  \textbf{d}).\textbf{b}: snapshots in time of the spin and temperature.\textbf{c}:  the convergence of the average local temperature to the known critical temperature of the 2D Ising model ($T_c \approx 2.269$) for a variety of initial temperatures. The global adaptation method was harder to balance and required more time to converge. \textbf{e}: A snapshot of spins after 25000 timesteps. The magnetization (\textbf{f}) and inverse temperature (\textbf{g}) are plotted over time.}}
\label{isingCApanel}
\end{figure*}

\begin{figure*}[!ht]
\centering
\includegraphics[width=\textwidth]{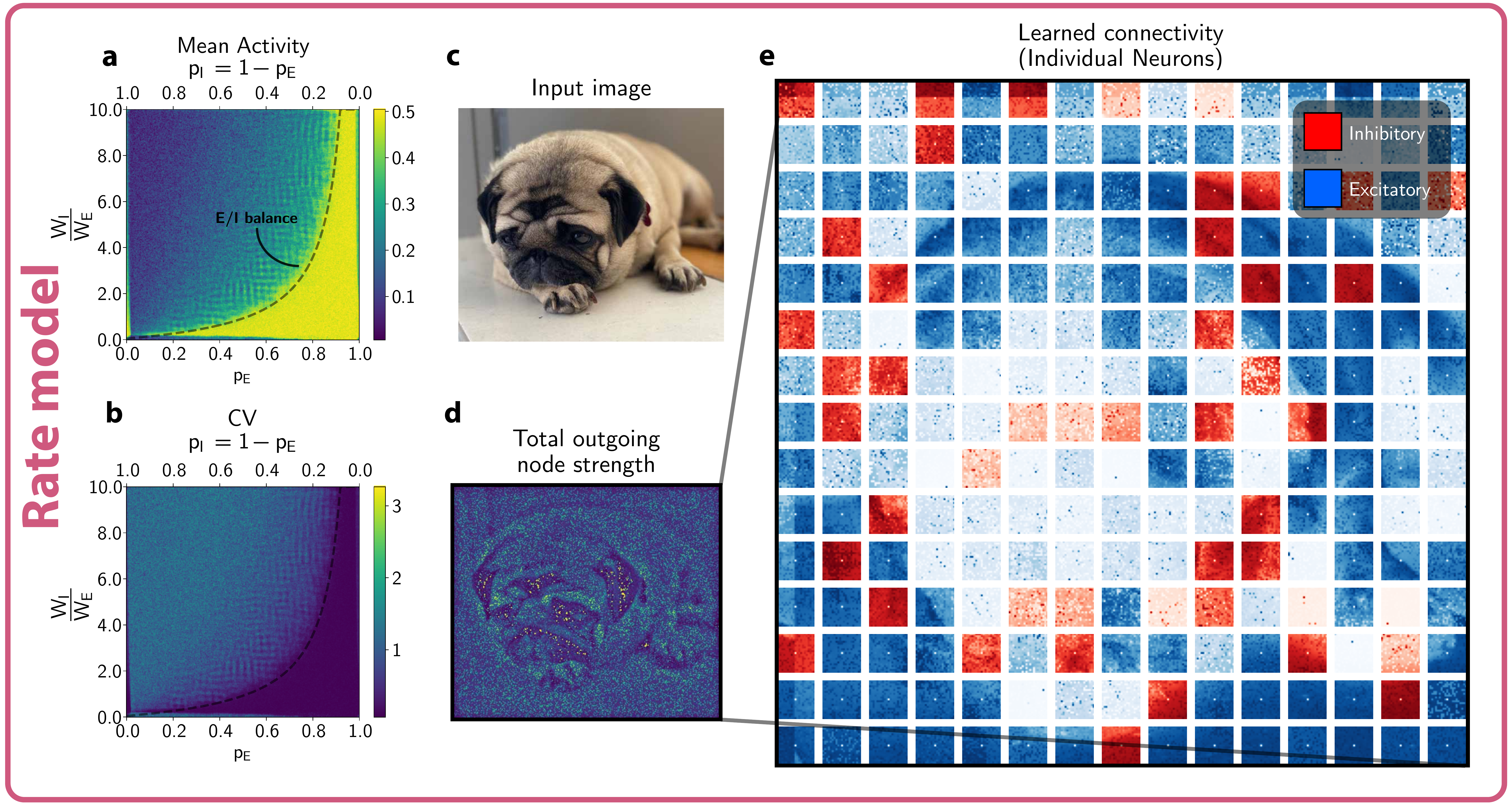}\vspace{-0.5em}
\caption{\textit{
(\textbf{a}): A phase diagram of the mean activity of the rate model as the excitatory neuronal density (probability of each neuron to be excitatory, $p_E$ is varied from $[0, 1]$  on the x-axis, and the relative weights of inhibitory/excitatory connections are varied from $[0, 10]$ on the y-axis. The dashed line marks the balance of average incoming excitatory and inhibitory connections to each neuron. (\textbf{b}): The coefficient of variation (std./mean) for the same range of parameters as (\textbf{a}). (\textbf{c}): A natural image is given as input to the network, and the connectivity after the convergence of plasticity (\textbf{d}). (\textbf{e}): The incoming weights matrices of 15x15 example neurons are shown, with inhibitory/excitatory neurons colored red/blue.
}}
\label{fig:ratenetwork}
\end{figure*}

\begin{figure*}[!ht]
\centering
\includegraphics[width=\textwidth]{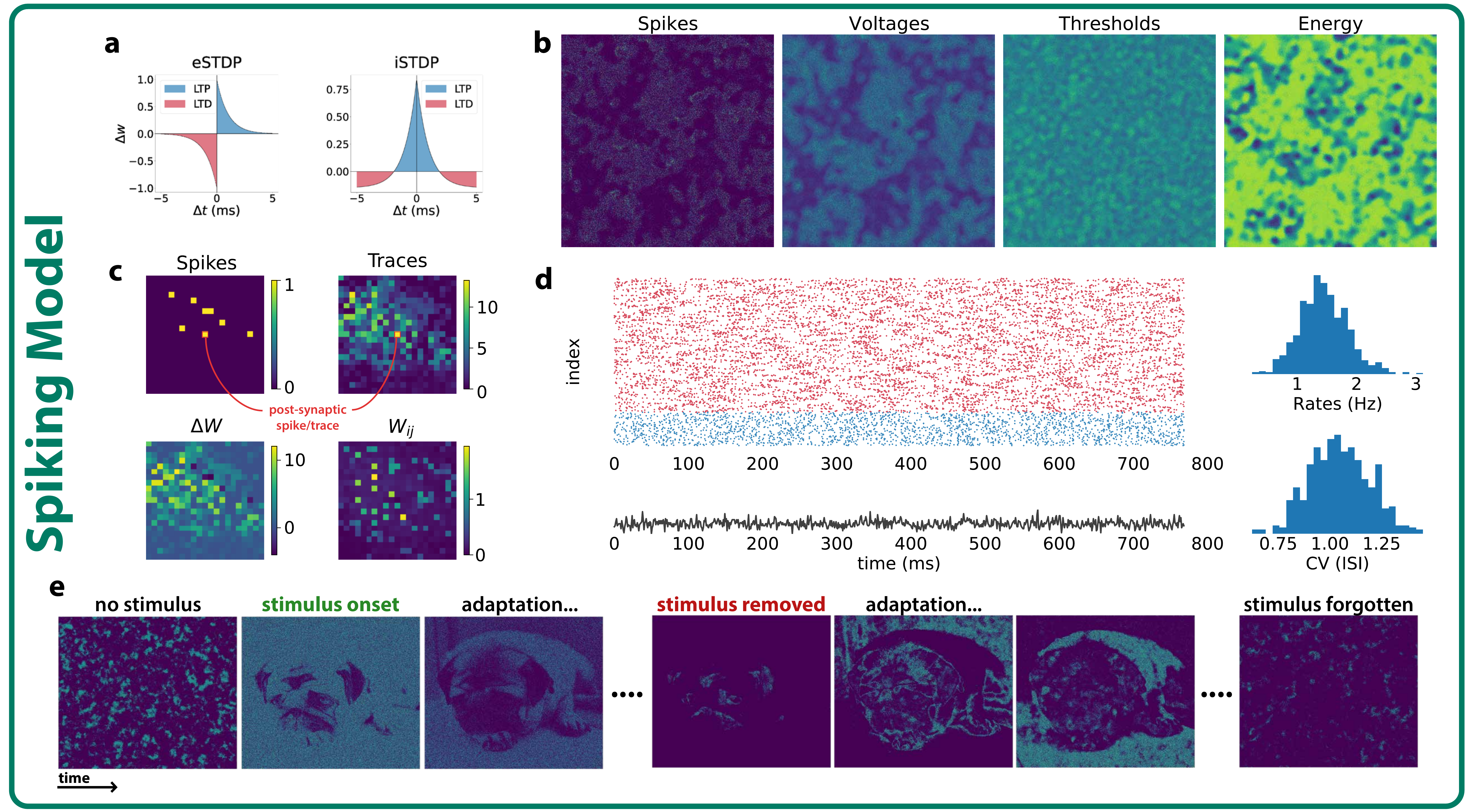}\vspace{-0.5em}

\caption{\textit{\textbf{a.} The synaptic plasticity rules for excitatory (left) and inhibitory (right) synapses. \textbf{b.} The activity of different channels of the CA. Each channel tracks different biophysical aspects of the neuron and is updated following different update rules. \textbf{c.} The pre-synaptic spikes and plasticity traces of a random neuron at location $i, j$ are shown, where the central pixel represents the post-synaptic neuron's spikes/traces. Applying the plasticity rule to this neuron, the connectivity is modified by $\Delta W$, and the resulting normalized connectivity $W_{ij}$ is shown in the bottom row. \textbf{d.} The network's spiking activity is asynchronous irregular (a widely observed state in biological neural networks), with a broad distribution of firing rates. The distribution of the CVs of the inter-spike intervals centered around $1$, indicating a Poisson-like activity. \textbf{e.} Threshold adaptation enables short-term memory in our model. Once the stimulus is presented, the activity quickly adapts to the statistics of the presented image. The inverted outline of the image persists in the network's activity for a short period after the stimulus is removed due to the relatively slow timescales of threshold adaptation.}}
\label{fig:spikingnetwork}
\end{figure*}

\section{Examples}

\subsection{Self-organized criticality in the Ising CA}
Phase transitions are fundamental phenomena in the study of complex systems as they demonstrate how matter changes its internal structure under varying conditions. Understanding the boundaries of different phases of organization and the conditions under which a system transitions from one phase to another is central to the understanding of both physical and living systems. Furthermore, there is increasing evidence that it is precisely at these boundaries between phases where many biological systems seem to operate, taking advantage of the computational/dynamical properties that characterize these critical points. However, setting a system to a critical point usually requires an accurate fine-tuning of the control parameters involved. Therefore, in order to claim that a biological system is at the edge of criticality, one has to show the existence of self-organizing mechanisms (SoC) that bring the control parameter to a critical point. This has been a common paradigm since the introduction of SOC models by \citep{bak1987self}. 

The Ising model has been, for the better part of a century, the foundational model to study the nature and mathematics behind second-order phase transitions \citep{ising1924beitrag}. Here we simulate a 2D Ising model CA and introduce mechanisms for the model to self-organize to criticality \citep{onsager1944crystal}. 

The Ising model has a set of state variables called spins $s_i$, which can be in the discrete positions of $-1$ or $+1$. The spins are connected in a 2D grid using a Von Neumann neighborhood connectivity pattern with common weights $J$. The system is characterized by its energy landscape, defined as:
\begin{equation}
    E\left(s\right) = - \sum_{\left\langle i, j \right\rangle} J_{ij} s_i s_j. 
\end{equation}
When $J>0$, spins tend to align to minimize their energy. In order to simulate the system, the Metropolis algorithm is used to update a randomly initialized state to arrive at an equilibrium configuration. An individual spin is flipped towards configurations that minimize its local energy. Flips that decrease the energy ($\Delta E < 0$) are always accepted, while the probability of energetically unfavorable ($\Delta E > 0$) flips are given by
\begin{equation}
    p(-s_{i}) = e^{-\Delta E_i \beta},
\end{equation}
where $\beta = 1/(kT)$, $k$ is the Boltzmann constant which we set to 1, and $T$ is the temperature of the model controlling the randomness of the system. At each timestep, a small fraction of the spins are allowed to update according to their flip probabilities. To make the system to self-organize to criticality, we allow the temperature to evolve depending on the system's state so that it can tune itself. Two methods, a local and a global method, are defined and applied to the model as it updates. 

The local adaptation method allows each cell/spin site to take a measurement of its nearest neighbors' magnetization
\begin{equation}
    m_i = \frac{1}{n} \sum_{j \in \langle nn \rangle}^n s_j
\end{equation}
and change its temperature according to an update rule that increases the local temperature if a system is too ordered (magnetized). The temperature is also allowed to diffuse, so neighboring cells will slowly average out their differences. The temperature decays at a rate proportional to itself. Combining all terms, growth, decay, and diffusion, we obtain:
\begin{equation}\label{eqn:ising_local_rule}
    \Delta T_i = \alpha m_i^2 - \epsilon T_i + D\nabla^2T_i,
\end{equation}
where $T_i\left(t + 1\right ) = \eta T_i\left(t \right) + \Delta T_i $. The values $\eta=\num{0.5}$, $\alpha=\num{1e-1}$, $\epsilon=\num{2e-2}$, $D = 1.$ were used for the local update rule. Pseudo-code of the patch operations required by this local model are summarized in Procedure \ref{alg:ising_soc}.

Alternatively, the global adaptation method updates the system's global temperature using global magnetization measurements:
\begin{equation}
    \Delta T = \alpha |M| - \xi(t)T^2, 
\end{equation}
where the global measurement of the system's magnetization is defined as:
\begin{equation}
    |M| = \left | \frac{1}{N} \sum_{j}^{N} s_j \right |.
\end{equation}
The update follows a similar rule as the local temperature logic: the ordered systems have their temperature increased, and the disordered systems -- decreased. A folded normal distribution scaled by $T^2$ is used to sample noise $\xi(t)$ for the decay term. The parameters of the global update rule share the same coefficients as the local method, with the exception of $\epsilon$, which is further divided by the linear size of the system. With a proper choice of parameters, both methods are capable of self-organizing the Ising model to its known critical temperature, see \fig\ref{isingCApanel}.

Some difficulties in discovering useful adaptation rules can be explained with the stability analysis of these coupled systems. In the examples we used, a homogeneous, linear stability analysis of a Hohenberg-Halperin model A (a macroscopic Langevin equation for the Ising model) combined with the adaptation equations shows that the critical fixed point might not always be stable, meaning that the mechanism is not truly self-organized. However, establishing a link between the parameters of the simulated model with Glauber dynamics and the macroscopic model is a difficult task, meaning that it is not possible to provide exact boundaries for sets of parameters that allow the model to self-organize. Despite this, we could find adequate parameter values simply by a trial-and-error approach, meaning that it is not complicated to make the model self-organized in practice.

Here, we demonstrate our SoC model on the 2D Ising network because the dynamics of this model are well-studied and drawn out before us, thus, we can verify if our methods perform as we expect them. However, one hopes to understand the broader landscapes of more complex and realistic systems, and that requires experimentation. Navigating the phase diagrams of arbitrary models in a goal-oriented way is likely to be challenging, so we explore these self-organizing models in order to understand via creation and experimentation how one might traverse these high-dimensional spaces with a principled approach.

\begin{algorithm}[!ht]
\caption{SOC 2D-Ising Model}\label{alg:ising_soc}
    \begin{algorithmic}[1]
        \Function{newTemp}{$spins, temps, \alpha, \epsilon, D$}
            \State $tPatches \gets$ \texttt{unfold($temps, patchSize$)}
            \State $sPatches \gets$ \texttt{unfold($spins, patchSize$)}
            \For{$T_{ij}, S_{ij}$ in \texttt{zip$(tPatches, sPatches)$}}
                \State $m_{ij} \gets$ \texttt{abs(mean($S_{ij}$))}
                \State $newT_{ij} \gets \alpha m_{ij}^2 - \epsilon T_{ij} + D \nabla^2 T_{ij}$
                \State $out_{ij} \gets $ \texttt{smooth}($T_{ij}, newT_{ij}$)
            \EndFor
            \State \textbf{return} $newT$
        \EndFunction
    \end{algorithmic}
\end{algorithm}

\subsection{Synaptic plasticity in a neural CA}
Simulating biological neural networks of a plausible scale in a reasonable amount of time is a well-established challenge in the field of computational neuroscience \citep{Giannakakis2020}. Given their ability to simulate millions of locally connected units in real time, CAs are a natural candidate for such large-scale simulations. Indeed, it has been recently shown that CAs can be fitted to model the activity of spiking neural networks \citep{Farner2021}. Nevertheless, one of the main features of biological neural networks is the ability of each individual neuron to learn and adapt to external stimuli by modifying its internal parameters and its connectivity to neighboring  neurons. This feature of biological neural networks is not possible to capture with traditional CAs, where the update rule of each pixel/neuron is the same as with all others and remains constant in time. The adaptive update rules can solve this problem by enabling the modeling of local threshold adaptation as well as synaptic plasticity in neural CAs. 

\subsubsection{Rate model:}
We first create a rate network consisting of neurons following  Wilson-Cowan dynamics \citep{Wilson1972} arranged in a 2D grid. Each neuron's activity is given by:
\begin{equation}
\tau \frac{dr}{dt} = -r + (1 - r)\cdot s(W R + c),
\end{equation}
where, $s$ is a sigmoid function, $W$ the neuron's connectivity matrix, $R$ the activity of neighboring neurons and $c$ a constant current.

Each neuron can be either excitatory or inhibitory and is connected to its neighboring neurons within a given radius that varies between simulations. 

We model distinct plasticity mechanisms active on the excitatory and inhibitory synapses (i.e., connections of the neuron follow a different plasticity rule depending on its type). Specifically, the weight of the connection from the excitatory neuron $i$ to neuron $j$ is modified according to a simple Hebbian rule \citep{hebb1949first}:

\begin{equation}
    \Delta W^E_{ij} = r_i \cdot r_j.
\end{equation}
While the connection weight from the inhibitory neuron $i$ to neuron $j$ is modified according to a homeostatic inhibitory rule \citep{Vogels2011inhibitory}:
\begin{equation}
    \Delta W^I_{ij} = r_i \cdot (r_j - \beta),
\end{equation}
where $\beta$ is a target activity level.
To further stabilize learning, we use synapse-type specific normalization that has been demonstrated to produce stable representations of inputs as well as E/I co-tuning in rate and spiking neural networks \citep{Eckmann2022, Giannakakis2023}.  The resulting network displays rich E/I dynamics, \fig\ref{fig:ratenetwork}a, b.

 Including this combination of plasticity mechanisms in neural CAs, allows us to study phenomena such as assembly formation \citep{Miehl2022} on a much larger scale than has ever been attempted before. In this study, we investigate whether natural images can be imprinted in the network's connectivity via synaptic plasticity. We project a natural image as additional incoming current $c$ each neuron receives (each neuron receives the grayscale value of a single pixel as input). Once the plasticity converges, we examine the learned connectivity. We plot (\fig \ref{fig:ratenetwork}d, e) the learned connectivity matrix of the network by coloring every neuron/pixel proportional to its out/in-going weights, respectively, showing that the input stimuli has been imprinted in the model's connectivity. Our results demonstrate that natural images can be consistently imprinted in our network's connectivity via a combination of simple plasticity mechanisms. 

\subsubsection{Spiking model:}
We then create a detailed simulation of a spiking neural network. We model different aspects of the neuron's behavior with different CA channels, each representing the local properties of a single cell and interacting with each other according to known differential equations.

The primary channel of the model tracks the membrane voltage of each neuron:
\begin{equation}
    \tau \frac{dV}{dt} = - V + I_e + c, 
\end{equation}
where $I_e$ is the input coming from other neurons and $c$ is a constant baseline current.
To demonstrate that we can use CAs to model neurons at an arbitrary level of biological accuracy, we use eLIF neurons (leaky integrate and fire neurons with an additional energy constraint) \citep{Fardet2020}, a model that demonstrates a high degree of biological plausibility, \fig\ref{fig:spikingnetwork} b. Thus, a neuron spikes when the voltage goes above a threshold $V_{th}$ and its energy is sufficiently high:
\begin{equation}
    S_t = \begin{cases}
        1, & \text{if  } \ V \geq V_{th}  \text{\ and  } \ E \geq E_{min}, \\
        0, & \text{otherwise.  }\\
    \end{cases} 
\end{equation}
where $S$ is the CA channel tracking spikes, $E$ is the neuron's energy and $E_{min}$ is the minimum energy required for spiking.

The energy $E$ is tracked by a different channel and follows the equation:
\begin{equation}
\tau_e \frac{dE}{dt} = \epsilon \cdot (E_0 - E) - s_c\cdot \sum_{f}\delta(t - t^f),
\end{equation}
where $\epsilon$ is an energy replenishment rate, $E_0$ the target energy, $s_c$ the energetic cost of a spike and $t_f$ the neuron's spike times ($t_f = \{ t: S_t = 1 \}$).

Moreover, each neuron's spiking threshold is modified by threshold adaptation, a homeostatic plasticity mechanism associated with increased robustness in spiking networks \citep{Fontaine2014, Huang2016}. Thus, a separate CA channel tracks the spike threshold, which changes according to the following equation:
\begin{equation}
\tau_{th} \frac{dV_{th}}{dt} = - V_{th} +  \eta_{th} \cdot (A - \rho_0),
\end{equation}
where $\eta_{th}$ is an adaptation rate, $\rho_0$ is a target firing rate, and $A$ is the neuron's spike trace, which is represented by a final CA channel as:
\begin{equation}
\tau_a \frac{dA}{dt} = - A + \sum_{f}\delta(t - t^f).
\end{equation}

The multiple CA channels allow an arbitrary degree of biological plausibility, with more details being potentially included by adding new channels. For example, more detailed voltage equations, refractoriness, and other details can be added if necessary.

We find that the spiking network exhibits realistic population dynamics. In particular, we see that it self-organizes in an asynchronous irregular firing state, a commonly observed dynamical regime in spiking networks \citep{Brunel2000}, see \fig\ref{fig:spikingnetwork} d. 

Additionally, we examine whether it can encode natural images similar to the simplified rate network. Indeed, we find that synaptic plasticity allows the encoding of images in the network's connectivity. Finally, the inclusion of threshold adaptation enables a short-term preservation of image statistics in the network's spiking activity once the stimulus has been removed. This observation is consistent with other findings about the relationship between short-term memory and threshold adaptation \citep{Itskov2011, Hu2021} and suggests that our model can replicate different key properties of spiking networks.

\subsection{Scalability and computational efficiency}
We tested the performance of our method compared to the standard neuronal activity simulator Brian2~\citep{stimberg2019brian}. On a single machine, our method allows simulating networks that are two orders of magnitude larger, where simulation time is scaling particularly favorably for large networks, \fig\ref{fig:benchmark}. Furthermore, the methods demonstrated in this paper are generalizable to a much broader variety of dynamical systems than most specialized and highly-optimized simulators of collective systems without too much loss of efficiency, can run and be visualized live, and can be made interactive with human intervention. We believe all these properties are essential for building deeper understandings of these emergent systems.

\section{Conclusion}

For the scientific study of the emergence and collective behavior in natural systems to become plausible in a systematic way, we first need to be able to reliably recreate and analyze such phenomena in simpler and well-understood artificial systems. Cellular automata are a natural candidate system for studying this kind of emergent behavior in a controlled and simplified manner. Here, we explore methods to expand the capabilities of CAs to model complex phenomena by introducing heterogeneity in their update rules.

Enabling CAs to be heterogeneous massively increases their expressivity. In fact, accurately describing any truly complex natural system requires some kind of heterogeneity in the modeling. Here, we demonstrate how the inclusion of such heterogeneity in the update rules of CAs can enable the modeling of complex natural systems in real-time on a very large scale. We use several examples of large-scale complex systems (Ising model, biological neural networks) that can utilize heterogeneous update rules to self-organize towards a desired state (critical point in the case of the Ising model, natural image derived attractor state in the case of the rate and spiking neural networks). As a follow-up example, one can study the self-organization of critical-like dynamics in neuronal networks with synaptic plasticities \citep{zeraati2020self}, or using the scalability of the system to previously impossible sizes to study the scaling behavior of the local networks \citep{zeraati2022topology}.

Our approach is not limited to the specific  systems we chose to model but is a general method that can be extended to many other complex, self-organized systems that require heterogeneous update rules. Any system with some kind of bottom-up organizational principle (i.e., locally interacting units) can potentially be modeled using locally adaptive CAs. Thus, we propose that adaptive CAs can be a useful tool to discover, either through experimentation or optimization, adaptation mechanisms that can self-organize dynamical systems in goal-oriented ways.
 \section{Acknowledgements}
This work was supported by a Sofja Kovalevskaja Award from the Alexander von Humboldt Foundation. EG and SK thank the International Max Planck Research School for Intelligent Systems (IMPRS-IS) for their support. We acknowledge the support from the BMBF through the T\"ubingen AI Center (FKZ: 01IS18039A). AL is a member of the Machine Learning Cluster of Excellence, EXC number 2064/1 – Project number 39072764. 
\footnotesize
\bibliographystyle{apalike}
\bibliography{main}

\end{document}